%% file: main.tex
\documentclass[10pt,twocolumn,letterpaper]{article}

\usepackage[pagenumbers]{cvpr} 

\input{preamble}

\definecolor{cvprblue}{rgb}{0.21,0.49,0.74}
\usepackage[pagebackref,breaklinks,colorlinks,allcolors=cvprblue]{hyperref}

\def\confName{CVPR}
\def\confYear{2027}

\title{FocusAnchor: Zero-Shot Metric Depth by Anchoring\\a Depth Foundation Prior to Focus Cues}

\author{
Ashkan Ganj$^{1}$ \quad Hang Su$^{2}$ \quad Tian Guo$^{1}$\\
$^{1}$Worcester Polytechnic Institute \quad $^{2}$NVIDIA Research\\
{\tt\small \{aganj,tian\}@wpi.edu, hangsu@nvidia.com}
}
\begin{document}
\raggedbottom
\maketitle
\input{fig/teaser}

\input{sec/0_abstract_v2}
\input{sec/1_intro_v2}
\input{sec/2_related}

\input{sec/3_method}
\input{sec/4_experiments}

\input{sec/5_conclusion_v2}
\input{sec/6_llm}
{
    \small
    \bibliographystyle{ieeenat_fullname}
    \bibliography{main}
}

\input{sec/X_suppl}

\end{document}

%% file: preamble.tex
\usepackage{cuted} 
\usepackage{currfile} 

\usepackage{caption} 

\usepackage{ragged2e}
\usepackage{colortbl}
\newcommand{\para}[1]{%
  \par\addvspace{3pt}%
  \noindent\textbf{#1}\ \ignorespaces
}

\usepackage{multirow}
\newcommand{\method}{FocusAnchor\xspace}
\newcommand{\best}[1]{\textbf{#1}}
\newcommand{\second}[1]{\underline{#1}}

%% file: fig/teaser.tex
\begin{strip}
\centering
\vspace{-25pt}

\includegraphics[width=\linewidth]{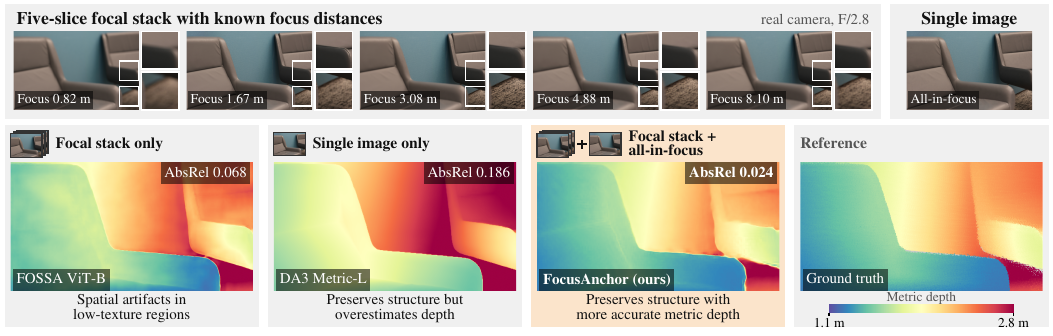}
\captionsetup{hypcap=false}
\captionof{figure}{\textbf{\method combines focus cues and a frozen depth prior for accurate metric depth.}
Top: a real five-slice focal stack from ZEDD~\cite{zuo2026fossa}, captured at known focus distances, and an all-in-focus image.
Bottom: metric depth predictions and ground truth. Numbers are per-image AbsRel, and all depth maps share the same color scale. In this example, FOSSA ViT-B~\cite{zuo2026fossa} exhibits spatial artifacts, while Depth Anything~3 Metric-Large~\cite{lin2025da3} overestimates depth.
By repeatedly anchoring the foundation prior to focus cues, \method improves metric accuracy while preserving scene structure.
}
\captionsetup{hypcap=true}
\label{fig:teaser}
\end{strip}

%% file: sec/0_abstract_v2.tex
\begin{abstract}
Existing depth-from-focus methods can struggle with weak focus cues, new cameras, or fewer focal slices. We present \method, a recurrent model that uses focus-volume lookups to refine metric depth and repeatedly aligns a relative depth prior from a frozen foundation model to the current depth estimate. The aligned prior guides subsequent refinement in a feedback loop, carrying metric scale recovered from textured regions into regions where focus cues are weak while preserving scene structure. To support transfer across cameras, we synthesize training stacks with randomized camera parameters and focus distances. However, stack synthesis blurs existing image noise differently across slices, creating artificial depth cues even on textureless surfaces. We add independent per-slice noise to suppress these cues. We also introduce variable-stack training to improve robustness with fewer focal slices and support more practical mobile capture. Compared with FOSSA ViT-B, \method reduces AbsRel by 36\% on NYU Depth V2 and by 9\% on DDFF-12 after fine-tuning. In zero-shot evaluation, \method and its variable-stack variant, \method-flex, achieve the two lowest AbsRel values on all five benchmarks. On ZEDD test, \method achieves 9\% lower AbsRel than the best competing method FOSSA ViT-B, while \method-flex achieves lower AbsRel with 40\% fewer focal slices on ZEDD validation. Check project page for code and weights at \url{https://cake.wpi.edu/FocusAnchor-website/}.
\end{abstract}

%% file: sec/1_intro_v2.tex
\vspace{-15pt}
\section{Introduction}

\label{sec:intro}

Metric depth estimation predicts the distance from a camera to visible surfaces in meters. The metric depths are useful because they enable applications such as augmented reality to correctly scale and position virtual objects~\cite{10.1145/3638550.3641122}, and help robots judge distances for obstacle avoidance and manipulation~\cite{9197518}. 
Camera focus can provide a metric cue because surfaces appear sharpest when focused at their distance.
Capturing the same scene at several known focus distances produces a focal stack, with each image called a \emph{focal slice}. Depth-from-focus (DFF) models~\cite{maximov2020defocusnet,wang2021aifdepthnet,yang2022dfv,woo2025dualfocus, ganj2025hybriddepth} estimate metric depth from sharpness changes across slices and their known focus distances. However, changes in sharpness are not equally informative across the image. Textured regions often provide informative focus cues, while smooth walls and other textureless surfaces can look similar across slices, making their depth difficult to determine.

Relative-depth foundation models such as Depth Anything V2~\cite{yang2024dav2} can provide complementary scene structure from a single image, including where focus cues are weak, but their predictions lack metric scale. HybridDepth~\cite{ganj2025hybriddepth} aligns a monocular depth estimate with a focus-based estimate and refines the result. FOSSA~\cite{zuo2026fossa} adapts a pretrained depth model to focal stacks, demonstrating zero-shot transfer to new cameras and scenes.

Despite these advances, three limitations remain. \emph{First}, metric depth can remain inaccurate in regions with weak focus cues, particularly textureless surfaces. \emph{Second}, achieving strong generalization with limited task-specific training data remains challenging. FOSSA, for example, is trained on stacks rendered from 373k synthetic images. \emph{Third}, existing methods often lose accuracy with fewer focal slices, limiting their use when capture time or scene motion restricts stack size. On an iPhone 18 Pro, we measure about 0.58\,s per slice and 2.9\,s for five slices (\cref{sec:stack}).

We introduce \method, a recurrent DFF model built around a feedback loop between depth refinement and prior alignment (\cref{fig:teaser}). Given a focal stack, its focus distances, and an all-in-focus image, \method maps a frozen foundation model's relative depth prediction onto the focus range. A recurrent network refines this estimate through lookups in a focus volume, which stores learned scores across focal slices at each pixel. After initial scale adjustment, each refinement step fits the frozen prediction to the current estimate using a confidence-weighted local scale and shift. The aligned prior and focus evidence guide the next update, carrying metric scale recovered from textured regions into regions where focus cues are weak.

To support cross-camera transfer, we train the refinement network under varied acquisition conditions using NYU Depth V2 images. Following FOSSA, we synthesize focal stacks with randomized camera parameters and focus distances. However, during the synthesis process, blurring a captured RGB image also smooths the sensor noise already present. At each pixel, this noise is smoothed least in the slice focused at the surface’s depth, creating an artificial depth cue even on textureless surfaces. A model can therefore learn to locate the in-focus slice from residual noise rather than scene detail. This shortcut does not reflect real capture, where sensor noise is introduced separately in each exposure rather than blurred with the scene. We therefore add independent random noise to each rendered slice to suppress this shortcut.
To support different capture budgets, we also train a variant with two to five focal slices, \method-flex.


\method achieves strong accuracy on held-out NYU Depth V2 images and on DDFF-12 after fine-tuning. Without target-dataset fine-tuning,  \method and \method-flex achieve the two lowest AbsRel values across five benchmarks, including real captured focal stacks in ZEDD, where \method reduces AbsRel by about 9\% relative to FOSSA ViT-B on the public test set. With variable-stack training, \method-flex maintains strong performance across stack sizes. On ZEDD validation, \method-flex achieves 73\% lower AbsRel than FOSSA ViT-B when both use two slices and achieves lower AbsRel with three slices than FOSSA ViT-B with five. These results suggest potential for mobile capture with fewer exposures.

Our main contributions are:
\begin{itemize}
\item A recurrent DFF architecture that updates depth in units of focus planes and repeatedly aligns a frozen depth prior to the current estimate to guide refinement.
\item A training recipe combining randomized camera parameters and independent per-slice noise to suppress artificial depth cues from synthetic stack rendering and support transfer to real captured stacks.
\item Robustness across focal-stack sizes through variable-stack training that enables \method-flex to support two to five slices and outperforms competing methods with two or three slices on ZEDD.
\item State-of-the-art zero-shot AbsRel on ZEDD, Infinigen Defocus, iBims, HAMMER, and ARKitScenes.
\end{itemize}

%% file: sec/2_related.tex
\section{Related Work}
\label{sec:related work}

\para{Depth from focus.}
Early DFF methods estimate depth using handcrafted sharpness measures across focal settings~\cite{suwajanakorn2015dff}. DDFF~\cite{hazirbas2018ddff} introduced end-to-end DFF and the DDFF-12 light-field benchmark, while DefocusNet~\cite{maximov2020defocusnet} narrowed the synthetic-to-real gap with a defocus-aware architecture and the FoD500 dataset.
DFV~\cite{yang2022dfv} uses a differential focus volume from neighboring-slice feature differences and estimates a distribution over focus distances, while DualFocus~\cite{woo2025dualfocus} adds spatial and focal variational constraints to distinguish depth boundaries from texture artifacts and to encourage physically consistent training.
We adapt DFV's focus volume for recurrent depth refinement.


\para{Single-image depth estimation.}
Relative depth models predict depth only up to an unknown scale and shift.
MiDaS~\cite{ranftl2020robustmonoculardepthestimation} and Depth Anything~\cite{yang2024da} generalize across domains through diverse training data, and Depth Anything V2~\cite{yang2024dav2} adds robustness and fine detail.
Marigold~\cite{ke2023repurposing} repurposes a pretrained diffusion model for zero-shot depth.
Metric depth models predict depth in meters from learned scene priors, often with known or estimated intrinsics.
ZoeDepth~\cite{bhat2023zoedepth} adds metric heads to relative-depth training.
Metric3D, Metric3D v2 and DA3 Metric~\cite{yin2023metric3d,hu2024metric3dv2,lin2025da3} predict depth in a canonical camera space and rescale predictions using the input focal length.
UniDepth and UniDepth V2~\cite{piccinelli2024unidepth,piccinelli2025unidepthv2} predict camera geometry alongside depth, and Depth Pro~\cite{bochkovskii2025depthpro} estimates the focal length, enabling metric prediction without supplied intrinsics.
However, the single-image models we evaluated still incur large metric errors on ZEDD (\cref{tab:zeroshot}).
We instead anchor a relative depth prior using focus observations at known distances.

\para{Fusing focus with single-image priors.}
HybridDepth~\cite{ganj2025hybriddepth} aligns a relative single-image depth map to a DFF estimate with a global scale and shift, then applies learned pixelwise scale corrections.
FOSSA~\cite{zuo2026fossa} integrates focus cues into a pretrained metric depth transformer through attention across slices and an embedding of each focus distance.
In stereo, MonSter~\cite{cheng2025monster} aligns a monocular prior inside a recurrent loop through learned per-pixel shift updates.
We instead align the prior in closed form with a local scale and shift before every step, and combine it with focus-volume lookups in units of focus planes.

\para{Recurrent refinement with depth priors.}
RAFT~\cite{teed2020raft} and RAFT-Stereo~\cite{lipson2021raftstereo} iteratively refine optical flow and disparity, respectively, using ConvGRUs guided by features retrieved from correlation volumes.
DEFOM-Stereo~\cite{jiang2025defom} initializes disparity from Depth Anything V2 and applies scale updates before additive refinement.Stereo Anywhere~\cite{bartolomei2025stereoanywhere} combines correlation volumes derived from stereo image features and monocular depth predictions.
We extend DEFOM-Stereo's recurrent backbone with focus-volume lookups and depth updates relative to known focus planes.

%% file: sec/3_method.tex
\input{fig/figure_architecture}

\section{Method}
\label{sec:method}

Given a focal stack $\{I_i\}_{i=1}^{N}$ with known focus distances $d_1 < \dots < d_N$ in meters and an all-in-focus image $I_\text{aif}$, we estimate a metric depth map $D$.
\Cref{fig:architecture} shows the architecture of \method, where a CNN encodes the focal stack into a \emph{focus volume} that scores the sharpness of each pixel in each slice, and a frozen Depth Anything V2 predicts the relative inverse depth from the all-in-focus image.
We map this prediction onto the focus range and recurrently correct its scale before refining it in focus-plane coordinates. During the latter stage, each iteration aligns the prior to the current estimate and uses the aligned prior together with focus-volume lookups to guide the next update.
Building on DEFOM-Stereo's recurrent backbone~\cite{jiang2025defom}, we introduce focus-plane lookups and updates (\cref{sec:update}), repeated prior alignment (\cref{sec:fusion}), and per-slice noise and variable-stack training (\cref{sec:data}).

\subsection{Primer}
\label{sec:primer}

\para{Focus volume.}
For a thin lens, a point at depth $D$ appears as a blur disk whose diameter grows with $|1/D - 1/d_i|$.
The point is thus sharpest in the slice focused nearest to it in inverse depth, so we estimate inverse depth throughout. We take the focus volume from DFV~\cite{yang2022dfv} (\cref{sec:encoders}).
The focus distances are known in meters, so the volume carries metric depth.
Textureless surfaces can have flat scores.

\para{Depth prior.}
Depth Anything V2~\cite{yang2024dav2} covers these surfaces. From one image, it predicts a relative inverse depth $z$ that captures the structure of the scene but matches the true inverse depth only up to an unknown scale and shift.

\para{RAFT-based update loop.}
We combine the volume with the prior through the update loop of RAFT-Stereo~\cite{teed2020raft,lipson2021raftstereo}, which refines an estimate at $1/4$ resolution over several iterations.
In each iteration, a \emph{lookup} reads the volume around the current estimate.
A ConvGRU takes the lookup, the estimate and fixed context features, and predicts a correction.
GRU weights are shared within each stage, and a learned convex upsampling gives the full-resolution output.
From DEFOM-Stereo~\cite{jiang2025defom} we keep two ideas: the loop starts from the single-image prior, and its first iterations rescale the estimate before the rest add corrections.

\subsection{Encoders}
\label{sec:encoders}
\label{sec:volume}
A shared CNN maps each slice $I_i$ to 32-channel features $F_i$ at $1/4$ resolution.
As in DFV, we pair each feature with its difference to the next slice, $[F_i,\, F_i - F_{i+1}]$ with $F_{N+1} = 0$, and apply three 3D convolutions along the stack, which gives the focus volume $C(i, p)$ over slices $i$ and pixels $p$ for any $N$.
Unlike DEFOM-Stereo, we add no Depth Anything features, so the focus volume stays independent of the prior.
A softmax over slices gives the volume-only depth $D_\text{vol} = \big(\sum_i \operatorname{softmax}_i(C) / d_i\big)^{-1}$, which we supervise as an auxiliary output.
The prior $z$ comes from a frozen Depth Anything V2 Large (ViT-L) applied to $I_\text{aif}$, which can be fused from the slices.
A trained copy of the first layers of its DPT head gives multi-scale features, and a context encoder maps them and $I_\text{aif}$ to the initial hidden state $h_0$ and the context features read by every iteration.

\subsection{Recurrent Refinement}
\label{sec:update}

\method uses focus-volume lookups and a three-level ConvGRU at $1/4$, $1/8$ and $1/16$ resolution (\cref{fig:architecture}, bottom).
The \emph{scale stage} corrects large initial scale errors through multiplicative updates, while the \emph{step stage} then refines each pixel with bounded updates in focus-plane coordinates. Each stage has its own GRU weights, shared by its iterations, and the hidden state carries over. 

\para{Plane index $u$.}
The focus planes sit at $\rho_i = d_1/d_i$, from $\rho_1 = 1$ down to $\rho_N$.
We estimate the normalized inverse depth $x = d_1/D$, so the nearest focus plane is at $x = 1$ and the metric depth is $D = d_1/x$.
We also express $x$ as a \emph{plane index} $u = g_\rho(x)$, which counts focus planes from the nearest one.
A pixel on the second plane has $u = 2$, and a pixel halfway between the second and third planes in inverse depth has $u = 2.5$. The map $g_\rho$ is piecewise linear with $g_\rho(\rho_i) = i$ and extends linearly beyond the end planes.
Stereo measures steps in pixels of disparity. We measure them in planes, so a step of $1$ moves the estimate to the next focus plane for every camera and focus schedule.

\para{Initialization $x_0$.}
The first estimate takes its shape from the prior and its scale from the focus planes.
We min-max normalize $z$ to $\hat z \in [0,1]$ and initialize $x_0 = \rho_N + (1 - \rho_N)\,\hat z$.
The nearest predicted surface thus lands on the nearest focus plane, and the farthest on the farthest plane.
Our focus distances are in meters, so $x_0$ starts at a metric scale. This scale is only approximate, because a scene rarely spans exactly the focus range.

\para{Scale stage ($t \le 4$).}
At each of the first four iterations, a scale lookup reads $C$ at $g_\rho(\sigma x_{t-1})$ for nine scale factors $\sigma \in [0.5, 2]$, with five taps around each.
It thus shows how well each rescaling of the current estimate agrees with the focus evidence.
From this lookup, the GRU predicts a per-pixel scale $s_t$, and the estimate becomes $x_t = s_t\,x_{t-1}$.

\para{Step stage ($t > 4$).}
In the remaining iterations, the lookup $L_t$ reads $C$ at $u_{t-1} + k\,\delta$ for $k \in \{-4, \dots, 4\}$, interpolating linearly between slices.
It thus covers $r = 4\delta$ planes on each side of the estimate, written $C(g_\rho(x_{t-1}) \pm r)$ in \cref{fig:architecture}.
Stereo pools its volume into coarser levels, which is not an option with five or fewer slices.
We instead read the same volume at two step sizes, $\delta = 0.5$ and $1$.
The lookup reads raw scores rather than a softmax, and samples outside the stack read zero.
From $L_t$ and the aligned prior (\cref{sec:fusion}), the GRU predicts a step along the plane index,
\begin{equation}
x_t = g_\rho^{-1}\!\big(u_{t-1} + \operatorname{clip}(\Delta u_t, -4, 4)\big),
\label{eq:delta}
\end{equation}
where the clip matches the span the lookup sees at $\delta = 1$.
Like RAFT-Stereo, we train with fewer iterations, $T = 4 + 8$, than we test with, $T = 4 + 12$.

\subsection{Prior Alignment}
\label{sec:fusion}
In the loop described so far, the prior enters only through $x_0$ and the context.
Where the volume is flat, as on a blank wall, the GRU can thus drift away from the prior's structure.
We therefore align $\hat z$ to the estimate before every step-stage iteration, and weight each pixel by how much the estimate can be trusted (\cref{fig:architecture}, bottom). The scale stage needs no alignment, since $x_0$ is itself an affine map of $\hat z$.

\para{Local fit.}
Since $\hat z$ is inverse depth up to scale and shift, we fit it to $x_{t-1}$ with a scale $a_p$ and a shift $b_p$ per pixel $p$.
We solve least squares over a $33 \times 33$ window $W(p)$ at $1/4$ resolution, about 130 full-resolution pixels wide, weighted by the previous confidence $c_{t-1}$ (with $c_4 = 1$).
Let $\mathbb{E}_p$ be the weighted mean over $W(p)$, pulled slightly toward the image-wide mean so that a window with no confident pixel falls back to the global fit (\cref{sec:supp_refit} of the supplementary material).
This gives:
\begin{equation}
a_p = \max\!\Big(0,\, \frac{\operatorname{Cov}_p[\hat z, x_{t-1}]}{\max(\operatorname{Var}_p[\hat z],\, \epsilon)}\Big), \quad
b_p = \mathbb{E}_p[x_{t-1}] - a_p\, \mathbb{E}_p[\hat z].
\end{equation}
with $\operatorname{Cov}_p$ and $\operatorname{Var}_p$ formed from $\mathbb{E}_p$ and $\epsilon = 10^{-6}$.
The aligned prior $\tilde{x}_t(p) = a_p\,\hat z(p) + b_p$ is clamped to $[\tfrac12 \min x_{t-1},\, 2 \max x_{t-1}]$, so it cannot run to near-zero inverse depth.
No gradient flows through the fit.

\para{Confidence.}
A small head $\psi$ predicts the probability $c_t$ that $x_{t-1}$ lies within about 5\% of the ground truth, from the finest hidden state $h_{t-1}$, the lookup $L_t$ and $\log x_{t-1} - \log \tilde{x}_t$.
Since $c_t$ depends on $\tilde{x}_t$, it weights the next fit.
We detach $c_t$ before the GRU, so $\psi$ learns only from $\mathcal{L}_c$.

\para{Fusion.}
The step-stage GRU receives $[x_{t-1}, \tilde{x}_t, c_t]$ in place of $x_{t-1}$ alone.
It can follow the aligned prior where the volume is flat and follow the volume where it is peaked.

\subsection{Training Loss}
\label{sec:training}
Convex upsampling brings each estimate to full resolution, giving the metric depths $D_1, \dots, D_T$ and the output $\hat D = D_T$.
We supervise all $T$ estimates, the volume-only depth and the confidence,
\begin{equation}
\begin{split}
\mathcal{L} = {} & \sum_{t=1}^{T} \gamma^{T-t} \big\| \log D_t - \log D^\ast \big\|_1 \\
& + \lambda_v \big\| \log D_\text{vol} - \log D^\ast \big\|_1 + \lambda_c \mathcal{L}_c,
\end{split}
\end{equation}
where $\|\cdot\|_1$ averages over valid pixels, $\gamma = 0.9^{15/(T-1)}$ as in RAFT-Stereo, $\lambda_v = 1$ and $\lambda_c = 0.1$.
The log makes the loss the same for depth and for inverse depth.
$\mathcal{L}_c$ is a binary cross-entropy averaged over the step-stage iterations.
We supervise $c_t$ with 
$\mathbf{1}[\,|\log D_{t-1} - \log D^\ast| < 0.05\,]$.

\subsection{Training}
\label{sec:data}
We train \method on NYU Depth V2~\cite{silberman2012nyu}, rendering focal stacks with FOSSA's render~\cite{zuo2026fossa}.
Camera parameters and focus distances vary per image to support cross-camera transfer (Suppl. \cref{sec:supp_render}).
\method-flex is the same architecture trained with variable stacks (\cref{sec:stack}).
We address the following two sources of artificial depth cues.

\para{Focus planes.}
Prior focal-stack methods~\cite{ganj2025hybriddepth,woo2025dualfocus} place the nearest and farthest planes at each image's minimum and maximum ground-truth depth, so the network only places pixels between two known depths (Suppl. \cref{fig:planes}).
Following FOSSA, 80\% of our stacks instead use random focus distances, with the nearest plane at 0.6 to 1\,m and the farthest 7 to 15 times farther.

\para{Per-slice noise.}
The renderer blurs one sharp photo once per focus distance, so the photo's sensor noise stays sharpest in the slice that is in focus.
On a textureless surface, such as a white wall, this noise is the only detail, so its sharpest slice hints at the depth. A real camera exposes every slice separately, with new noise each time, so we add fresh Gaussian noise to every rendered slice.
If textureless pixels carried no depth hint, their sharpest slice would be the one focused nearest the true depth 20\% of the time, one in five.
In our renders this happens 62\% of the time, and the added noise lowers it to 27\% (Suppl. \cref{sec:supp_render}).

%% file: fig/figure_architecture.tex
\begin{figure}[t]
  \centering
  \includegraphics{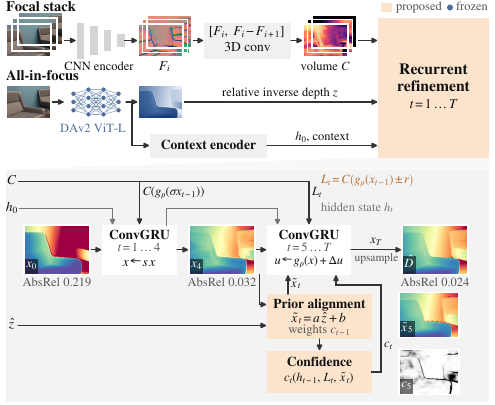}
    \caption{\textbf{FocusAnchor.}
    Top: a CNN encoder and 3D convolutions turn the focal stack into the focus volume $C$, and a frozen Depth Anything V2 gives the relative inverse depth $z$ of the AiF image.
    The recurrent refinement turns these into $x_T$, which is upsampled to the metric depth $\hat D = d_1/x_T$.
    Bottom: the refinement detail.
    The scale stage ($t \le 4$) rescales the estimate. The step stage ($t > 4$) reads the focus volume around the estimate ($L_t$) and moves it along the plane index $u$.
    Before each step, $\hat z$ is aligned to the estimate with weights $c_{t-1}$, which gives the aligned prior $\tilde x_t$. Confidence $c_t$ (white: high confidence) weights the next prior fit.}
  \label{fig:architecture}
  \vspace{-5pt}
\end{figure}

%% file: sec/4_experiments.tex
\section{Experiments}
\label{sec:experiments}

We evaluate three aspects of \method: (i) generalization across scene and camera domains, (ii) sensitivity to focal-stack size, and (iii) the contributions of prior alignment and iterative refinement. Trained only on NYU Depth V2, \method and \method-flex achieve the two lowest AbsRel values on all five zero-shot benchmarks (\cref{tab:zeroshot}).
\method-flex remains accurate with as few as two slices, which suggests practical utility when a long focal stack cannot be captured.

\subsection{Experimental Setup}
\label{sec:setup}

\input{sec/4_experiments_setup_datasets_v2}

\para{Metrics.}
Following prior work~\cite{zuo2026fossa,ganj2025hybriddepth,woo2025dualfocus}, we report the AbsRel, SqRel, RMSE, logRMSE and threshold accuracy $\delta_t$, the fraction of pixels with $\max(\hat d_p/d_p,\, d_p/\hat d_p) < t$.
$\delta_1$, $\delta_2$ and $\delta_3$ denote $t = 1.25$, $1.25^2$ and $1.25^3$.
For DDFF-12, the metrics are computed on disparity, and we also report MSE. For the ZEDD test split, we use the official evaluation server.


\para{Implementation details.}
We train \method on the data of \cref{sec:data} for 60k steps on $384 \times 512$ crops with five slices, using AdamW with a learning rate of $2\times10^{-4}$, a one-cycle schedule, and a batch of 8 on two RTX 4090 GPUs.
\method-flex is trained likewise with 2--5 slices per batch, then for 20k more steps at a learning rate of $5\times10^{-5}$.
Following FOSSA~\cite{zuo2026fossa}, we use the ZEDD validation split during development: \method uses its final 60k-step checkpoint, while \method-flex uses a checkpoint selected on this split. Each model is evaluated using a single fixed checkpoint across all zero-shot benchmarks.
For comparison with HybridDepth and DualFocus on NYU, we separately train \method using their rendering protocol: the NYU camera, focus planes spanning each image's ground-truth depth range (Suppl.~\cref{fig:planes}, top), and no per-slice noise augmentation. For DDFF-12, we fine-tune on its training split.
\method has 24.6M trainable parameters plus the frozen Depth Anything V2 Large model. Five-slice $480\times640$ inference takes 92\,ms on an RTX 4090.

\input{tabs/table1_nyu}
\input{tabs/table2_ddff}

\subsection{Performance on NYU Depth V2 and DDFF-12}
\label{sec:indomain}

\para{NYU Depth V2.}
HybridDepth and DualFocus train with the same render as the NYU test stacks: the NYU camera and focus planes fixed to each image's ground-truth depth range.
As this makes a direct comparison unfair, we also train an in-domain version of \method with their render, without per-slice noise, and compare it with them.
In \cref{tab:nyu}, our default model reduces AbsRel by 29\% compared with the best single-image model and 36\% compared with FOSSA ViT-B. The in-domain version lowers the AbsRel of DualFocus by 41\% and its RMSE by 25\%.

\para{DDFF-12.}
Against the earlier focus methods, fine-tuned \method lowers AbsRel by 38\% and RMSE by 24\% relative to DualFocus (\cref{tab:ddff}).
Against FOSSA, the gap closes.
Our AbsRel is 9\% below that of FOSSA ViT-B, and the RMSE is tied (0.5\%, below the precision FOSSA reports).
FOSSA ViT-B keeps the lead on MSE, SqRel, $\delta_2$ and $\delta_3$, and FOSSA ViT-S on $\delta_1$.
A likely reason is that fine-tuning gives every model the target camera, which removes the camera transfer where our design gains most.

\input{tabs/table3_zeroshot}
\input{fig/fig_qual}

\subsection{Zero-Shot Generalization Performance}
\label{subsec:zero_shot}

We show that \method and \method-flex, trained only on NYU Depth V2, achieve the two lowest AbsRel values on all five benchmarks, which differ in scenes and cameras (\cref{tab:zeroshot}). Qualitative results are shown in \cref{fig:qual}.

\para{ZEDD.}
On real captured stacks, \method and \method-flex rank first and second on test AbsRel and $\delta_{1.05}$ (\cref{tab:zeroshot}). 
Compared with FOSSA ViT-B, the strongest competing method on this split, \method reduces AbsRel by 9\% and raises $\delta_{1.05}$ from 0.505 to 0.580.
FOSSA ViT-B retains the highest test $\delta_{1.25}$ (0.918 versus 0.903), so our advantage is clearest at the stricter threshold.
On validation, both variants rank in the top two across all three metrics, with 24\% lower AbsRel than FOSSA ViT-B.
Both variants preserve scene structure with lower per-image AbsRel than the baselines in all ZEDD and ARKitScenes examples in \cref{fig:qual}.

\para{Infinigen Defocus.}
Both variants occupy the top two positions on all reported metrics (\cref{tab:zeroshot}), extending the gains to rendered scenes. Relative to FOSSA ViT-S, \method-flex lowers AbsRel by 16\%, while \method raises \(\delta_{1.05}\) by 23\%. \method-flex also lowers AbsRel by 94\% relative to HybridDepth and 88\% relative to DFV.

\para{iBims and HAMMER.}
On iBims, under FOSSA's synthetic-stack protocol (\cref{tab:zeroshot}),
\method reduces AbsRel by 50\% compared to FOSSA ViT-B (0.070 to 0.035) and increases $\delta_{1.25}$ from 0.963 to 0.986.
On HAMMER, \method further reduces AbsRel by 41\% relative to FOSSA ViT-B (0.017 to 0.010).

\para{ARKitScenes.}
Both variants rank in the top two on all three metrics (\cref{tab:zeroshot}).
\method lowers AbsRel by 27\% relative to FOSSA ViT-B and places 4\% more pixels within 5\% of the ground truth.
Both methods exceed 0.99 in $\delta_{1.25}$. 
In the ARKitScenes examples in \cref{fig:qual}, we recover more accurate metric scale on close-up walls and ceilings. We achieve these gains using only NYU Depth V2 for training, whereas FOSSA uses the larger Hypersim and TartanAir datasets.

\input{tabs/table4_stack}

\subsection{Robustness Across Focal-Stack Sizes}
\label{sec:stack}

We evaluate zero-shot robustness to focal-stack size on the 50 ZEDD validation scenes (\cref{tab:stack}). For each scene, we use slices $\{1,9\}$, $\{1,5,9\}$, $\{1,4,6,9\}$ or $\{1,3,5,7,9\}$ of its nine focus settings, and fuse $I_\text{aif}$ from these slices only (Suppl.~\cref{sec:supp_stack}).
All methods in \cref{tab:stack} were trained with five slices, except \method-flex, which was trained with stack sizes sampled from two to five (\cref{sec:setup}).

\para{Robustness with variable-stack training.}
\method-flex is less sensitive to reductions in input stack size while retaining comparable five-slice accuracy. Reducing the stack from five to two slices increases its AbsRel by only 1.6X, compared with 4.3X/4.5X/2.4X for \method, FOSSA ViT-B, and HybridDepth, respectively. It achieves the best results among the compared methods on all three metrics with two to four slices.
With two slices, its AbsRel is 73\% lower than FOSSA ViT-B and 68\% lower than HybridDepth.
At five slices, it nearly matches \method's AbsRel with small tradeoffs in RMSE and $\delta_{1.25}$.

\para{Practical capture and inference implications.}
Acquiring a focal stack requires repeated focus adjustments and photo captures.
On iOS, processing to improve photo quality can further delay image delivery~\cite{apple2025avfoundation}.
Longer acquisition windows also make it harder to maintain a consistent viewpoint and scene during handheld capture.
We quantify this cost using an iOS app on an iPhone 18 Pro that locks the lens at five evenly spaced focus positions and captures each photo once the lens has settled. Across 10 trials, a five-slice stack takes 2.89\,s on average, and photo capture is 85\% of the 579\,ms per slice (supplementary \cref{tab:timing}).
Reducing the stack to two slices thus shortens acquisition by 60\%, about 1.7\,s, under this protocol.
Together with its stronger accuracy than the compared methods at two slices (\cref{tab:stack}), \method-flex offers a practical tradeoff between depth accuracy and acquisition time, allowing fewer focal slices at inference when capture time is limited.

\input{sec/4_experiments_alations_v2}

%% file: sec/4_experiments_setup_datasets_v2.tex
\para{Datasets.} Our evaluation covers captured, light-field-refocused, and synthetic focal stacks. 

\begin{itemize}
\setlength{\itemsep}{2pt}

\item \textbf{NYU Depth V2}~\cite{silberman2012nyu} contains indoor RGB-D images, with 24,231 training and 654 test images in our split. We evaluate on the test images following prior work~\cite{ganj2025hybriddepth,woo2025dualfocus}, with focus planes spanning each image's ground-truth depth range.

\item \textbf{DDFF-12}~\cite{hazirbas2018ddff} provides focal stacks refocused from light-field images of 12 indoor scenes. We use five-slice inputs and follow prior protocols~\cite{hazirbas2018ddff,yang2022dfv} for comparison.

\item \textbf{ZEDD}~\cite{zuo2026fossa} provides focal stacks captured with a full-frame camera and a 50\,mm lens, with dense LiDAR ground-truth. We evaluate its validation and test splits, each containing 50 scenes. We follow FOSSA evaluation and test protocols. Inputs are resized and center-cropped to $480\times640$, with valid ground-truth depths restricted to $[0.01,1000]$\,m.

\item \textbf{Infinigen Defocus}~\cite{zuo2026fossa} is a synthetic benchmark containing 200 indoor scenes generated with Infinigen Indoors~\cite{10655244} and rendered with defocus in Blender Cycles. Following FOSSA, we use five images per scene at F/1.4, with focus distances of 0.8, 1.7, 3.0, 4.7, and 8.0\,m, retaining ground-truth depths in $[0.01,1000]$\,m. For \method, RGB images and depth maps are resized to $480\times640$ using Lanczos and nearest-neighbor interpolation, respectively.

\item \textbf{iBims, HAMMER, and ARKitScenes} provide RGB-D images without focal stacks. For zero-shot evaluation, we synthesize focal stacks from their RGB images and ground-truth depth maps using the rendering procedure from FOSSA's evaluation~\cite{zuo2026fossa}. For ARKitScenes, we use 5,594 validation images downsampled to $480\times640$.

\end{itemize}

%% file: tabs/table1_nyu.tex
\begin{table}[t]
\centering
\caption{Metric depth estimation results on \textbf{NYU Depth V2}. \method achieves the lowest AbsRel and RMSE within each training group.}
\label{tab:nyu}
\setlength{\tabcolsep}{3pt}
\resizebox{\linewidth}{!}{%
\begin{tabular}{@{}lccccccc@{}}
\toprule
Method & RMSE$\downarrow$ & logRMSE$\downarrow$ & AbsRel$\downarrow$ & SqRel$\downarrow$ & $\delta_1\uparrow$ & $\delta_2\uparrow$ & $\delta_3\uparrow$ \\
\midrule
\multicolumn{8}{@{}l}{\emph{NYU rendering protocol}} \\
HybridDepth$^\dagger$~\cite{ganj2025hybriddepth} & 0.128 & -- & 0.026 & -- & \second{0.995} & \best{1.000} & \best{1.000} \\
DualFocus$^\dagger$~\cite{woo2025dualfocus} & \second{0.075} & \second{0.020} & \second{0.013} & \second{0.0021} & \best{0.999} & \best{1.000} & \best{1.000} \\
\method{} & \best{0.0564} & \best{0.0157} & \best{0.0077} & \best{0.0012} & \best{0.999} & \best{1.000} & \best{1.000} \\
\midrule
\multicolumn{8}{@{}l}{\emph{Default training protocols}} \\
UniDepth V2~\cite{piccinelli2025unidepthv2} & \second{0.297} & \second{0.101} & 0.080 & 0.041 & 0.950 & \second{0.991} & \second{0.995} \\
DA3 Metric-L~\cite{lin2025da3} & 0.309 & 0.104 & 0.075 & \second{0.036} & 0.961 & 0.989 & \second{0.995} \\
Metric3D v2~\cite{hu2024metric3dv2} & 0.317 & 0.207 & \second{0.072} & 0.038 & \best{0.965} & 0.989 & 0.993 \\
FOSSA ViT-S~\cite{zuo2026fossa} & 0.458 & 0.240 & 0.094 & 0.080 & 0.884 & 0.945 & 0.965 \\
FOSSA ViT-B~\cite{zuo2026fossa} & 0.403 & 0.209 & 0.080 & 0.061 & 0.906 & 0.958 & 0.975 \\
\cmidrule{1-8}
\method & \best{0.1874} & \best{0.0713} & \best{0.0513} & \best{0.0172} & \second{0.962} & \best{0.999} & \best{1.000} \\
\bottomrule
\end{tabular}}
\begin{minipage}{\linewidth}
\scriptsize\justifying
\noindent  \emph{NYU rendering protocol}: training uses the NYU camera and focus planes spanning each image's ground-truth depth range. \emph{Default training}: each method uses its own training settings and \method follows \cref{sec:data}.
Results are from our evaluations using released checkpoints for prior methods, except $^\dagger$ taken from the original papers.
Best and second-best results within each group are bold and underlined.
\end{minipage}
\vspace{-5pt}
\end{table}

%% file: tabs/table2_ddff.tex
\begin{table}[t]
\centering
\caption{Depth estimation results on \textbf{DDFF-12}. \method achieves the lowest AbsRel and RMSE.}
\label{tab:ddff}
\setlength{\tabcolsep}{2.5pt}
\resizebox{\linewidth}{!}{%
\begin{tabular}{@{}lccccccc@{}}
\toprule
Method & MSE$\downarrow$ & RMSE$\downarrow$ & AbsRel$\downarrow$ & SqRel$\downarrow$ & $\delta_1\uparrow$ & $\delta_2\uparrow$ & $\delta_3\uparrow$ \\
\midrule
DFV~\cite{yang2022dfv} & 5.7e-4 & 0.0213 & 0.17 & 0.0063 & 0.767 & 0.942 & 0.981 \\
HybridDepth~\cite{ganj2025hybriddepth} & 5.1e-4 & 0.0200 & 0.17 & 0.0060 & 0.789 & 0.947 & 0.981 \\
DualFocus~\cite{woo2025dualfocus} & 4.7e-4 & 0.0194 & 0.16 & 0.0056 & 0.800 & 0.954 & 0.982 \\
FOSSA ViT-S~\cite{zuo2026fossa} & 4.2e-4 & 0.0183 & \second{0.11} & 0.0045 & \best{0.936} & \second{0.983} & \second{0.991} \\
FOSSA ViT-B~\cite{zuo2026fossa} & \best{2.8e-4} & \second{0.0148} & \second{0.11} & \best{0.0025} & \second{0.932} & \best{0.987} & \best{0.994} \\
\midrule
\method & \second{3.0e-4} & \best{0.01472} & \best{0.10} & 0.0035 & \second{0.932} & 0.980 & 0.990 \\
\bottomrule
\end{tabular}}
\par\vspace{3pt}
\begin{minipage}{\linewidth}
\scriptsize\justifying
\noindent \method is fine-tuned on the DDFF-12 training split using five-slice stacks and evaluated at the final checkpoint. Metrics are computed on disparity. Best and second-best results are bold and underlined, respectively.
\end{minipage}
\end{table}

%% file: tabs/table3_zeroshot.tex
\providecommand{\testval}[2]{}
\renewcommand{\testval}[2]{#1\,/\,{\scriptsize #2}}

\begin{table*}[t]
\centering

\caption{\textbf{Zero-shot metric depth estimation across five benchmarks.}  \method and \method-flex outperform all evaluated single-image and depth-from-focus baselines on AbsRel, achieving the top two positions on every benchmark.}
\label{tab:zeroshot}
\setlength{\tabcolsep}{2pt}
\resizebox{\linewidth}{!}{%
\begin{tabular}{@{}lccc|ccc|ccc|ccc|ccc@{}}
\toprule
& \multicolumn{3}{c|}{ZEDD (Test / Val)}
& \multicolumn{3}{c|}{Infinigen Defocus}
& \multicolumn{3}{c|}{iBims~\cite{koch2018evaluationcnnbasedsingleimagedepth}}
& \multicolumn{3}{c|}{HAMMER~\cite{jung2022my}}
& \multicolumn{3}{c}{ARKitScenes~\cite{baruch2021arkitscenes}} \\
Method
& $\delta_{1.05}\uparrow$ & $\delta_{1.25}\uparrow$ & AbsRel$\downarrow$
& $\delta_{1.05}\uparrow$ & $\delta_{1.25}\uparrow$ & AbsRel$\downarrow$
& $\delta_{1.05}\uparrow$ & $\delta_{1.25}\uparrow$ & AbsRel$\downarrow$
& $\delta_{1.05}\uparrow$ & $\delta_{1.25}\uparrow$ & AbsRel$\downarrow$
& $\delta_{1.05}\uparrow$ & $\delta_{1.25}\uparrow$ & AbsRel$\downarrow$ \\
\midrule
\multicolumn{16}{c}{\emph{Single image}} \\
\midrule
DAv2~\cite{yang2024dav2}
& \testval{0.037}{0.044}
& \testval{0.261}{0.253}
& \testval{0.337}{0.338}
& 0.026 & 0.133 & 0.429
& 0.268 & 0.885 & 0.127
& 0.043 & 0.322 & 0.541
& 0.088 & 0.440 & 0.394 \\
Depth Pro~\cite{bochkovskii2025depthpro}
& \testval{0.182}{0.148}
& \testval{0.665}{0.615}
& \testval{0.201}{0.219}
& 0.174 & 0.692 & 0.176
& 0.308 & 0.878 & 0.120
& 0.200 & 0.720 & 0.271
& 0.019 & 0.093 & 1490.2 \\ 
UniDepth V2~\cite{piccinelli2025unidepthv2}
& \testval{0.106}{0.146}
& \testval{0.605}{0.589}
& \testval{0.253}{0.348}
& 0.103 & 0.464 & 0.370
& 0.510 & 0.942 & 0.078
& 0.068 & 0.482 & 0.371
& 0.296 & 0.832 & 0.133 \\
MoGe-2~\cite{wang2025moge2}
& \testval{0.149}{0.137}
& \testval{0.580}{0.520}
& \testval{0.278}{0.309}
& 0.175 & 0.621 & 0.233
& 0.394 & 0.890 & 0.106
& 0.300 & 0.751 & 0.269
& 0.422 & 0.879 & 0.096 \\
Metric3D v2$^\ast$~\cite{hu2024metric3dv2}
& \testval{0.007}{0.043}
& \testval{0.120}{0.326}
& \testval{0.558}{0.428}
& 0.074 & 0.374 & 0.401
& 0.188 & 0.761 & 0.184
& 0.039 & 0.463 & 0.346
& 0.239 & 0.749 & 0.160 \\
DA3 Metric-L$^\ast$~\cite{lin2025da3}
& \testval{0.202}{0.135}
& \testval{0.617}{0.684}
& \testval{0.230}{0.217}
& 0.204 & 0.707 & 0.186
& 0.471 & 0.954 & 0.076
& 0.168 & 0.716 & 0.218
& 0.351 & 0.890 & 0.101 \\
\midrule
\multicolumn{16}{c}{\emph{Depth from focus}} \\
\midrule
FV~\cite{yang2022dfv}
& \testval{0.153}{0.219}
& \testval{0.576}{0.759}
& \testval{0.369}{0.235}
& 0.045 & 0.172 & 0.629
& 0.231 & 0.759 & 0.141
& 0.560 & 0.912 & 0.106
& 0.361 & 0.935 & 0.103 \\
DFV~\cite{yang2022dfv}
& \testval{0.152}{0.227}
& \testval{0.546}{0.809}
& \testval{0.573}{0.148}
& 0.053 & 0.205 & 0.568
& 0.229 & 0.756 & 0.144
& 0.529 & \second{0.927} & 0.076
& 0.365 & 0.946 & 0.086 \\
DEReD~\cite{si2023fully}
& \testval{0.128}{0.125}
& \testval{0.505}{0.442}
& \testval{0.347}{0.389}
& 0.108 & 0.446 & 0.355
& 0.064 & 0.275 & 0.689
& 0.000 & 0.000 & 5.313
& 0.013 & 0.062 & 2.640 \\
HybridDepth~\cite{ganj2025hybriddepth}
& \testval{0.075}{0.270}
& \testval{0.347}{0.771}
& \testval{0.754}{0.153}
& 0.037 & 0.157 & 1.232
& 0.222 & 0.795 & 0.146
& 0.450 & 0.908 & 0.085
& 0.398 & 0.952 & 0.082 \\
FOSSA ViT-S~\cite{zuo2026fossa}
& \testval{0.451}{0.445}
& \testval{0.884}{0.886}
& \testval{0.098}{0.0985}
& 0.520 & 0.940 & 0.0847
& 0.519 & 0.954 & 0.075
& 0.601 & \best{0.999} & 0.044
& 0.780 & 0.995 & 0.034 \\
FOSSA ViT-B~\cite{zuo2026fossa}
& \testval{0.505}{0.432}
& \testval{\best{0.918}}{0.919}
& \testval{0.089}{0.0957}
& 0.420 & 0.936 & 0.0908
& 0.555 & 0.963 & 0.070
& 0.938 & \best{0.999} & 0.017
& 0.904 & \second{0.997} & 0.022 \\
\midrule
\method
& \testval{\best{0.580}}{\second{0.548}}
& \testval{\second{0.903}}{\best{0.927}}
& \testval{\best{0.081}}{\best{0.0731}}
& \best{0.639} & \second{0.941} & \second{0.0717}
& \best{0.799} & \best{0.986} & \best{0.035}
& \second{0.967} & \best{0.999} & \best{0.010}
& \best{0.943} & \best{0.998} & \best{0.016} \\
\method-flex
& \testval{\second{0.550}}{\best{0.570}}
& \testval{0.896}{\second{0.920}}
& \testval{\second{0.086}}{\second{0.0732}}
& \second{0.611} & \best{0.948} & \best{0.0708}
& \second{0.764} & \second{0.983} & \second{0.039}
& \best{0.970} & \best{0.999} & \second{0.012}
& \second{0.922} & \best{0.998} & \second{0.020} \\
\bottomrule
\end{tabular}}
\par\vspace{3pt}
\begin{minipage}{\linewidth}
\scriptsize\justifying
\noindent ZEDD entries report test/validation results, ranked separately for each split. ZEDD test and Infinigen results are from FOSSA~\cite{zuo2026fossa}, except for $\ast$ methods and our models, which we evaluate ourselves. All ZEDD validation and ARKitScenes results are from our evaluations. Best and second-best results are bold and underlined, respectively.
\end{minipage}
\vspace{-0.5cm}
\end{table*}

%% file: fig/fig_qual.tex
\begin{figure*}[t]
\centering
\includegraphics[width=\linewidth]{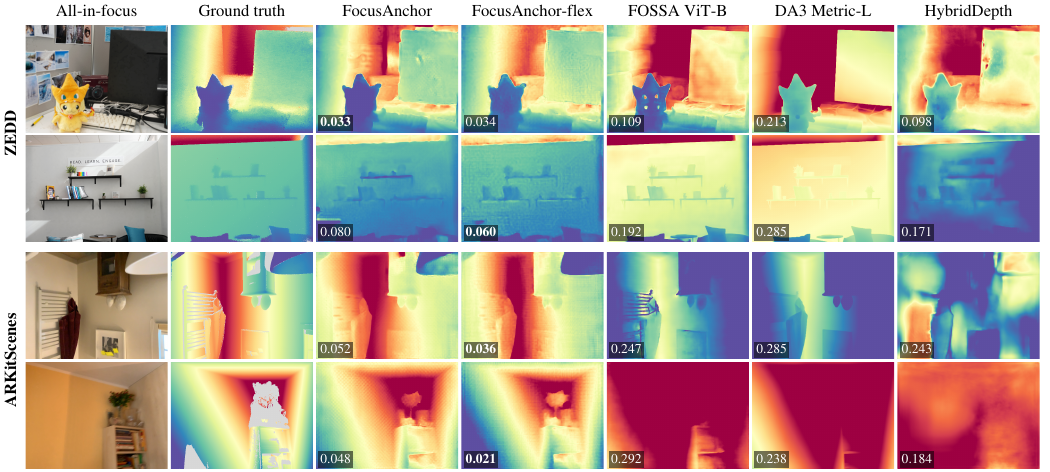}
\caption{\textbf{Zero-shot qualitative results} on ZEDD (top) and ARKitScenes (bottom).
Numbers are per-image AbsRel, with the lowest in each row in bold.
Per-pixel error maps and more visual results are in the suppl. \cref{fig:qual_error} and \cref{sec:supp_qual}.
}
\label{fig:qual}
\vspace{-3pt}
\end{figure*}

%% file: tabs/table4_stack.tex
\begin{table}[t]
\centering
\caption{\textbf{Zero-shot robustness to stack size on ZEDD validation (50 scenes).}
\method-flex leads on all metrics with two to four slices
and retains comparable five-slice accuracy.}
\label{tab:stack}
\footnotesize
\setlength{\tabcolsep}{1.5pt}
\renewcommand{\arraystretch}{1.0}
\begin{tabular*}{\linewidth}
{@{\extracolsep{\fill}}l*{6}{r}@{}}
\cmidrule[\heavyrulewidth](l){2-7}
& \multicolumn{3}{c}{2 slices}
& \multicolumn{3}{c}{3 slices} \\
\cmidrule(lr){2-4}
\cmidrule(l){5-7}
& AbsRel$\downarrow$ & RMSE$\downarrow$
& $\delta_{1.25}\uparrow$
& AbsRel$\downarrow$ & RMSE$\downarrow$
& $\delta_{1.25}\uparrow$ \\
\midrule
HybridDepth~\cite{ganj2025hybriddepth}
& 0.374 & 2.011 & 0.306
& 0.267 & 1.550 & 0.518 \\

FOSSA ViT-B~\cite{zuo2026fossa}
& 0.434 & 1.916 & \underline{0.421}
& \underline{0.206} & \underline{0.949}
& \underline{0.680} \\

\method
& \underline{0.313} & \underline{1.685} & 0.308
& 0.244 & 1.293 & 0.547 \\

\method-flex
& \textbf{0.119} & \textbf{0.774} & \textbf{0.855}
& \textbf{0.090} & \textbf{0.643} & \textbf{0.899} \\
\midrule
& \multicolumn{3}{c}{4 slices}
& \multicolumn{3}{c}{5 slices} \\
\cmidrule(lr){2-4}
\cmidrule(l){5-7}
& AbsRel$\downarrow$ & RMSE$\downarrow$
& $\delta_{1.25}\uparrow$
& AbsRel$\downarrow$ & RMSE$\downarrow$
& $\delta_{1.25}\uparrow$ \\
\midrule
HybridDepth~\cite{ganj2025hybriddepth}
& 0.231 & 1.249 & 0.607
& 0.153 & 1.021 & 0.771 \\

FOSSA ViT-B~\cite{zuo2026fossa}
& \underline{0.134} & \underline{0.756}
& \underline{0.857}
& \underline{0.096} & 0.722 & 0.919 \\

\method
& 0.158 & 1.015 & 0.755
& \textbf{0.073} & \textbf{0.588} & \textbf{0.927} \\

\method-flex
& \textbf{0.112} & \textbf{0.726} & \textbf{0.876}
& \textbf{0.073} & \underline{0.603}
& \underline{0.920} \\
\bottomrule
\end{tabular*}
\par\vspace{3pt}
\begin{minipage}{\linewidth}
\scriptsize
All models use five-slice training while \method-flex additionally uses two to five slices.
Bold/underlined are best/second-best per slice count and metric. 
\end{minipage}
\vspace{-5pt}

\end{table}

%% file: sec/4_experiments_alations_v2.tex
\subsection{Ablations}
\label{sec:ablations}

\input{tabs/table5_ablation}

\input{tabs/table6_ablation_retrained}

We ablate \method on ZEDD validation, NYU Depth V2, and ARKitScenes through test-time modifications (\cref{tab:ablation}) and variants retrained under a common 10k-step schedule (\cref{tab:ablation_retrained}). The results support combining focus cues with repeated prior alignment and highlight the importance of training-stack synthesis for cross-camera transfer.

\para{Complementary cues (\cref{tab:ablation}).}
The full model outperforms either single-cue estimate on all three datasets. Among component removals, disabling focus lookups causes the largest degradation on NYU and ARKitScenes, while removing the scale stage has the largest effect on ZEDD. Removing the fitted-prior input increases AbsRel by 17-24\%, confirming its consistent contribution.

\para{Prior alignment.}
\cref{tab:ablation} shows that fitting the prior only once after the scale stage, while retaining it as a GRU input, increases AbsRel by approximately 4-10\% relative to the full model. This suggests that updating the prior's alignment as the depth estimate changes provides better guidance than keeping the initial fit fixed. Local fitting and learned confidence have small effects overall. In the retrained ablations (\cref{tab:ablation_retrained}), removing prior alignment or the scale stage also increases AbsRel on all three datasets.

\para{Training-stack synthesis (\cref{tab:ablation_retrained}).}
Training with the NYU evaluation camera configuration improves NYU accuracy but reduces transfer to other cameras: ZEDD AbsRel roughly doubles, and ARKitScenes AbsRel increases by 76\%. Removing per-slice noise worsens ZEDD accuracy with both randomized and fixed cameras. On NYU, noise removal improves accuracy with the fixed camera but worsens it with randomized cameras.

\para{Refinement iterations.}
With five slices, the four scale iterations provide most of the accuracy gain. Reducing subsequent step iterations from twelve to eight keeps AbsRel within 1.3\% of the default for both models across all three datasets and reduces inference time by 10\%. Two-slice \method-flex benefits from longer refinement on ZEDD, suggesting the accuracy-runtime tradeoff depends on stack size. Detailed results are in \cref{sec:supp_iterations} of the supplementary material.

%% file: tabs/table5_ablation.tex
\begin{table}[t]
\centering
\caption{\textbf{Test-time ablations of \method (AbsRel $\downarrow$).}
The full model outperforms either single-cue estimate on all
three datasets. Removing focus lookups causes the largest
degradation on NYU and ARKitScenes, while removing the
scale stage has the largest effect on ZEDD.}
\label{tab:ablation}
\footnotesize
\setlength{\tabcolsep}{3pt}
\renewcommand{\arraystretch}{1.0}
\begin{tabular*}{\linewidth}{@{\extracolsep{\fill}}lrrr@{}}
\cmidrule[\heavyrulewidth](l){2-4}
& ZEDD & NYU & ARKitScenes \\
\midrule
\textbf{\method{} (full)}
    & \textbf{0.0731} & \textbf{0.0513} & \textbf{0.0163} \\
\midrule
\multicolumn{4}{@{}l}{\textit{Single-cue estimates}} \\
Prior only ($x_0$)
    & 0.485\phantom{0} & 0.0962 & 0.0883 \\
Focus volume only ($D_{\mathrm{vol}}$)
    & 0.141\phantom{0} & 0.0576 & 0.0326 \\
\midrule
\multicolumn{4}{@{}l}{\textit{Test-time modifications}} \\
w/o focus lookup (step stage)
    & \cellcolor{black!8}0.113\phantom{0}
    & \cellcolor{black!18}0.107\phantom{0}
    & \cellcolor{black!18}0.0900 \\
w/o scale stage
    & \cellcolor{black!18}0.115\phantom{0}
    & 0.0598
    & \cellcolor{black!8}0.0386 \\
w/o fitted-prior input
    & 0.0857
    & \cellcolor{black!8}0.0602
    & 0.0202 \\
Single prior fit ($t=5$)
    & 0.0797 & 0.0566 & 0.0170 \\
Global prior fit
    & 0.0752 & \textbf{0.0513} & 0.0173 \\
Uniform confidence ($c=1$)
    & \textbf{0.0731} & 0.0516 & 0.0164 \\
\bottomrule
\end{tabular*}
\par\vspace{3pt}
\begin{minipage}{\linewidth}
\scriptsize
All variants use the same 60k-step checkpoint without
retraining and the ZEDD validation split.
Bold denotes the lowest AbsRel, while darker and lighter gray mark the largest and second-largest increases in AbsRel relative to the full model among test-time modifications, respectively, within each dataset.
The prior-only estimate $x_0$ is mapped to the focus range.
\end{minipage}
\vspace{-0.5cm}

\end{table}

%% file: tabs/table6_ablation_retrained.tex
\begin{table}[t]
\centering
\caption{\textbf{Retrained ablations of \method (AbsRel $\downarrow$).}
Camera randomization is important for generalization, as the fixed-camera variants incur the largest errors on ZEDD and ARKitScenes. Removing prior alignment or the
scale stage also degrades accuracy on all three datasets.}
\label{tab:ablation_retrained}
\footnotesize
\setlength{\tabcolsep}{3pt}
\renewcommand{\arraystretch}{1.0}
\begin{tabular*}{\linewidth}{@{\extracolsep{\fill}}lrrr@{}}
\cmidrule[\heavyrulewidth](l){2-4}
& ZEDD & NYU & ARKitScenes \\
\midrule
\textbf{\method{} (full)}
    & \textbf{0.0813} & 0.0447 & \textbf{0.0233} \\
\midrule
\multicolumn{4}{@{}l}{\textit{Architecture}} \\
w/o prior alignment
    & 0.0876
    & \cellcolor{black!8}0.0553
    & 0.0264 \\
w/o scale stage
    & 0.0914 & 0.0520 & 0.0254 \\
\midrule
\multicolumn{4}{@{}l}{\textit{Training configuration}} \\
w/o per-slice noise
    & 0.0914
    & \cellcolor{black!18}0.0558
    & 0.0236 \\
Fixed camera$^{*}$
    & \cellcolor{black!8}0.1622
    & 0.0153
    & \cellcolor{black!8}0.0411 \\
Fixed camera, no noise$^{*}$
    & \cellcolor{black!18}0.2474
    & \textbf{0.0134}
    & \cellcolor{black!18}0.0417 \\
\bottomrule
\end{tabular*}
\par\vspace{3pt}
\begin{minipage}{\linewidth}
\scriptsize
All variants are only trained for 10k steps.
Bold denotes the lowest AbsRel, while darker and lighter gray mark the largest and second-largest increases in AbsRel relative to the full model, respectively, within each dataset.
$^{*}$ The training camera matches the NYU evaluation renderer, making these NYU results in-domain.
\end{minipage}
\vspace{-0.5cm}
\end{table}

%% file: sec/5_conclusion_v2.tex
\section{Conclusion}
\label{sec:conclusion}

\method uses an aligned foundation-model depth prior to carry metric scale recovered from textured regions into regions where focus cues are weak. \method and \method-flex achieve competitive zero-shot performance, and ablations demonstrate the benefits of repeatedly aligning the prior with focus-derived depth estimates throughout refinement. Our variable-stack training supports shorter focal stacks, but does not explicitly account for camera or scene motion during acquisition. Future work could incorporate these effects into model design and training to improve robustness in handheld capture.

%% file: sec/6_llm.tex
\section*{LLM Usage Statement}
We used large language models (LLMs) to refine the writing of this paper, mainly to improve the clarity, grammar, and phrasing of text drafted by the authors.
The research ideas, method design, choice of datasets, implementation, and experiments are the authors' own work, and LLMs were not used to generate research content.
The authors reviewed all text and take full responsibility for the content of the paper.

%% file: sec/X_suppl.tex
\clearpage
\setcounter{page}{1}
\maketitlesupplementary

\section{Prior Alignment Details}
\label{sec:supp_refit}
For the local fit in \cref{sec:fusion}, we compute confidence-weighted averages in each window $W(p)$ using the confidence map $c = c_{t-1}$.
Let $\mu_p[\cdot]$ denote the mean over $W(p)$ and $\langle \cdot \rangle$ the mean over the full image.
At image borders, $W(p)$ includes only pixels inside the image.

For each $m \in \{\hat z,\, x_{t-1},\, \hat z^2,\, \hat z x_{t-1}\}$, we use a pseudo-count $\kappa = 0.05$ to pull the local average toward the image-wide confidence-weighted average.
The resulting average is
\begin{equation}
\mathbb{E}_p[m] =
\frac{\mu_p[c\,m] + \kappa\,\langle c\,m\rangle / \langle c\rangle}
     {\mu_p[c] + \kappa}.
\end{equation}
If all pixels in $W(p)$ have zero confidence, each average equals its image-wide value, so the window uses the global fit.

We use these averages to compute the local variance and covariance:
\begin{align*}
\operatorname{Var}_p[\hat z]
&= \mathbb{E}_p[\hat z^2] - \mathbb{E}_p[\hat z]^2, \\
\operatorname{Cov}_p[\hat z, x_{t-1}]
&= \mathbb{E}_p[\hat z\,x_{t-1}]
 - \mathbb{E}_p[\hat z]\,\mathbb{E}_p[x_{t-1}].
\end{align*}

\section{Rendering Training Focal Stacks}
\label{sec:supp_render}

We use FOSSA's thin-lens renderer~\cite{zuo2026fossa} to generate $384 \times 512$ focal stacks from NYU Depth V2 with a 36\,mm sensor.
For each image, we scale the focal length by a zoom factor drawn from $[1, 1.5]$ and select an f-number from $\{1.0, 1.4, 2.0, 2.8, 4.0\}$.
We use a $15 \times 15$ power-exponential blur kernel with exponent $2^u$, where $u \sim \mathcal{U}(1,5)$.
The kernel is Gaussian at exponent 2 and approaches a disk as the exponent increases to 32.

\begin{figure}[t]
\centering
\includegraphics[width=\linewidth]{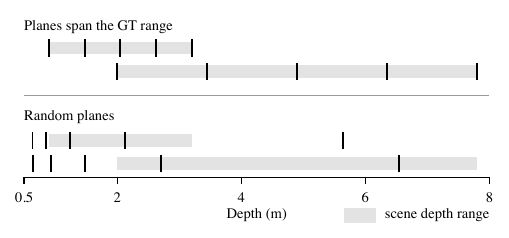}
\caption{\textbf{Focus planes during training.} Gray bars show the ground-truth depth range of each image. Prior work~\cite{ganj2025hybriddepth,woo2025dualfocus} spaces the planes across this range. Following FOSSA~\cite{zuo2026fossa}, we sample focus distances independently of the image's depth range for 80\% of training images.}
\label{fig:planes}
\end{figure}

\para{Focus planes.}
For 80\% of training images, we sample the nearest focus distance $d_1 \sim \mathcal{U}(0.6, 1)$\,m and set the farthest distance to $d_N = r d_1$, where $r \sim \mathcal{U}(7, 15)$.
We draw $\beta$ uniformly from $(0, 1]$ and set the $N$ focus distances to
\begin{equation}
d_i = \Big((1-\tau_i)\,d_1^{-\beta} + \tau_i\,d_N^{-\beta}\Big)^{-1/\beta},
\qquad \tau_i = \frac{i-1}{N-1}.
\end{equation}
When $\beta = 1$, the planes are evenly spaced in inverse depth. As $\beta \to 0$, their spacing becomes uniform in log depth.
For the remaining 20\% of images, we space the planes evenly in inverse depth between the 5th and 95th percentiles of each image's inverse depth (\cref{fig:planes}).

FOSSA samples the f-number, blur kernel, and focus planes once per batch. We use the same sampling scheme and draw a new set of parameters for each image at every training step.

\para{Per-slice noise.}
Every rendered slice gets its own Gaussian noise, drawn independently of the other slices.
Its standard deviation is drawn per image from $\mathcal{U}(0, 0.01)$, for intensities in $[0, 1]$.

\para{Noise test.}
A pixel is textureless when the Sobel gradient magnitude of the gray all-in-focus image, averaged over a $7 \times 7$ window, is below $0.0265$ for intensities in $[0, 1]$.
This covers 21\% of the valid pixels of the training render.
The sharpest slice of a pixel is the one with the largest absolute Laplacian, averaged over $3 \times 3$.
The test passes when this slice is the one focused nearest to the ground truth in inverse depth.
We average the pass rate per image over 300 NYU training images.
Without per-slice noise, textureless pixels pass 62\% of the time, against 20\% by chance with five slices.
With the noise, they pass 27\%.
This likely stays above chance because some of these pixels hold faint real texture, which is a true focus cue.
Textureless pixels in real ZEDD stacks pass the same test 41\% of the time.

\section{Number of Refinement Iterations}
\label{sec:supp_iterations}

\begin{figure}[t]
\centering
\includegraphics[width=\linewidth]{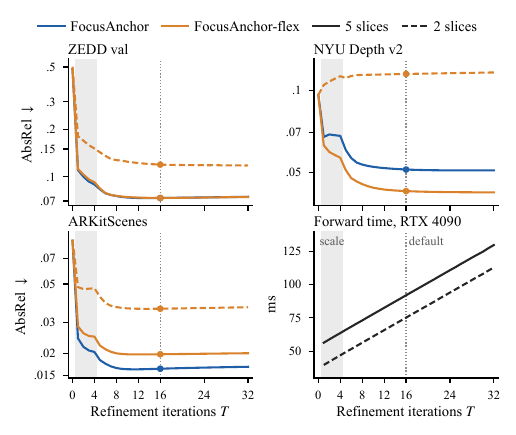}
\caption{\textbf{Accuracy and speed against the number of refinement iterations.}
AbsRel$\downarrow$ after $T$ iterations, where $T=0$ is the start $x_0$, and the forward time for $T$ iterations.
The shaded band marks the four scale iterations. The dotted line and the dots mark the default $T=16$ ($4+12$) used in all tables.
\method and \method-flex share one architecture, so their times coincide with five slices.
Times are medians of 100 passes on ZEDD val at $480\times640$, batch 1, bfloat16, on one RTX 4090.
On NYU the focus planes span the ground-truth depth range (\cref{sec:setup}), so with two slices the start already has the true range.}
\label{fig:iterations}
\end{figure}

\Cref{fig:iterations} shows AbsRel after each iteration of a single run of up to 32 iterations. Since each update depends only on earlier states, the state at iteration $T$ is also the output we would obtain by stopping at $T$.
The first four iterations correct scale errors in the initialization, which maps the prior's minimum and maximum values to the ends of the focus range.
On ZEDD with five slices, the first scale iteration reduces AbsRel for \method from 0.485 at $x_0$ to 0.110. After all four scale iterations, AbsRel reaches 0.089, an 82\% reduction from $x_0$.
The step stage reduces AbsRel by a further 18\% to 0.073, with little change after about eight step iterations.
\Cref{fig:supp_refine} shows how the depth and error maps change during refinement for three ZEDD scenes.

With five slices, stopping after four scale iterations and eight step iterations ($4+8$, as in training) saves 10\% of the forward-pass time. AbsRel stays within 1.3\% of the 16-iteration default for both models on all three datasets.
Stopping at $4+4$ saves 21\% of the time and increases AbsRel by up to 6\%.
Beyond 16 iterations, the effect varies by dataset. On NYU, AbsRel improves by at most 1\%. On ZEDD and ARKitScenes, it rises by 1--3\% at 32 iterations. One possible reason is that the GRU was trained for only 12 iterations.

Each iteration takes 2.36\,ms. With five slices, the default 16 iterations account for 41\% of the 92\,ms forward pass. The frozen DAv2 ViT-L backbone, the slice encoder, and the focus volume account for the remaining 54\,ms. This explains why halving the iteration count saves about one fifth of the total time.

With two slices, \method-flex converges more slowly on ZEDD. At $4+8$, its AbsRel is 2.3\% above the default and continues to fall slightly through iteration 32.
On NYU, refinement increases two-slice AbsRel by 19\% relative to $x_0$. The initialization already uses the ground-truth depth range, and the two focus planes lie at its endpoints. These planes may provide too little focus information to improve the initial estimate.

\begin{figure*}[t]
\centering
\includegraphics[width=\linewidth]{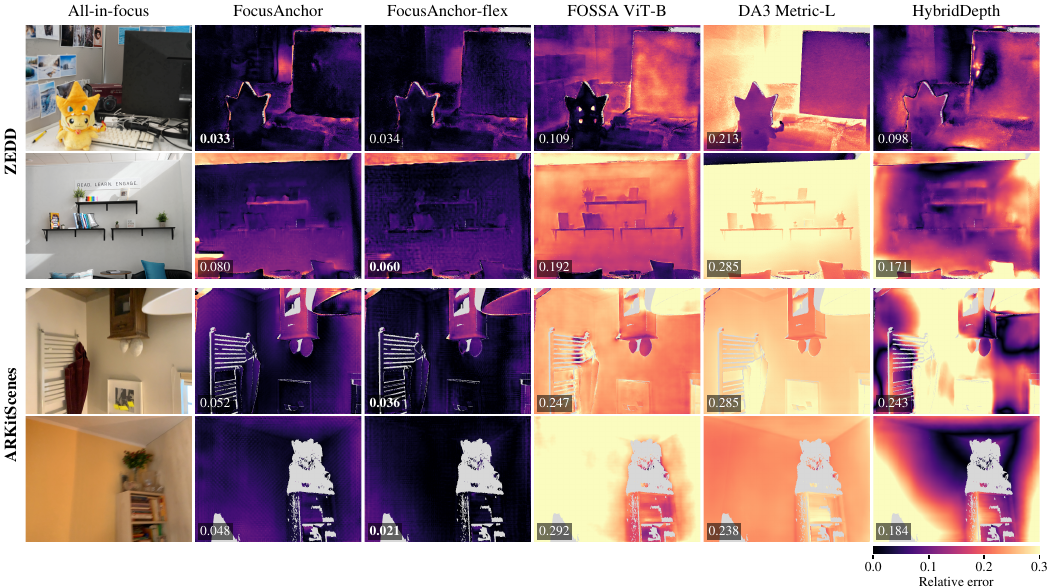}
\caption{\textbf{Per-pixel relative error} for the scenes of \cref{fig:qual}, clipped at 0.3.
Gray marks pixels without ground truth.
Numbers are per-image AbsRel, with the lowest in each row in bold.
FOSSA and Depth Anything~3 err almost evenly across each scene, because they misjudge its metric scale.
The errors of \method and \method-flex sit mainly at depth edges.}
\label{fig:qual_error}
\end{figure*}

\begin{figure*}[p]
\centering
\includegraphics[width=\linewidth]{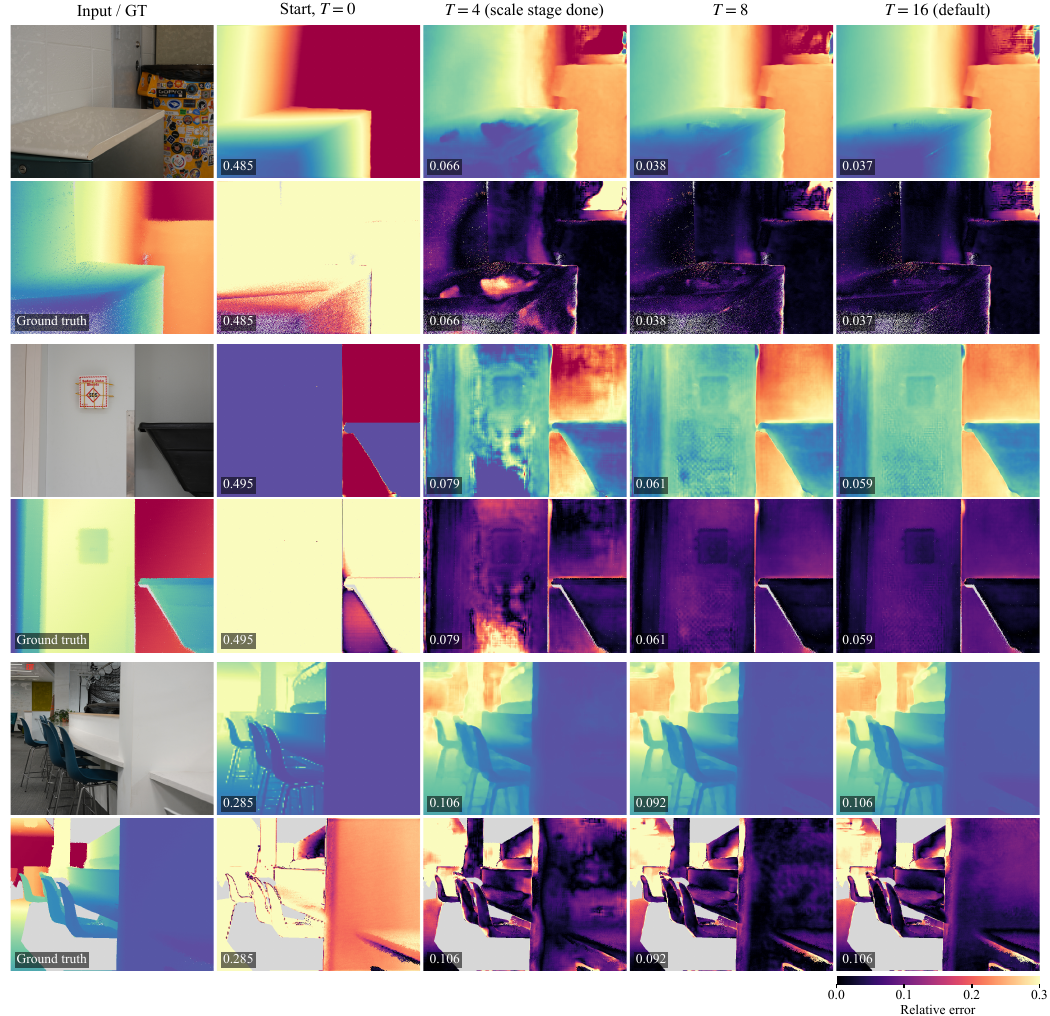}
\caption{\textbf{Depth and error during refinement.} For each ZEDD scene of \cref{fig:supp_qual_a}, the top row shows the depth of \method after $T$ iterations and the bottom row its relative error. The first column holds the all-in-focus image and the ground truth. Numbers are per-image AbsRel.}
\label{fig:supp_refine}
\end{figure*}

\section{Focal-Stack Capture Time on a Phone}
\label{sec:supp_timing}

\begin{table}[t]
\centering
\caption{\textbf{Focal-stack capture time on a phone.} iPhone 18 Pro, standard photo API, 10 five-slice stacks (50 slices). Every photo was exposed after the lens had settled.}
\label{tab:timing}
\setlength{\tabcolsep}{6pt}
\begin{tabular}{@{}lcc@{}}
\toprule
Step & Mean $\pm$ std & Range \\
\midrule
Lens move until settled & 87 $\pm$ 11\,ms & 71--106\,ms \\
Photo capture & 491 $\pm$ 8\,ms & 478--510\,ms \\
\midrule
One slice & 579 $\pm$ 13\,ms & 555--609\,ms \\
Five-slice stack & 2.89 $\pm$ 0.03\,s & 2.85--2.96\,s \\
\bottomrule
\end{tabular}
\end{table}

\Cref{tab:timing} breaks down the time to capture a focal stack on a phone.
Our iOS app moves the lens to five evenly spaced positions across its focus range.
At each position, it waits until the lens has settled and then requests a photo.
Photo capture takes 85\% of each slice, and the lens move takes the remaining 15\%.
Within capture, the shutter fires about 0.1\,s after the request, and the photo is delivered about 0.39\,s later.
Because every slice costs about the same, stack time grows linearly with the number of slices.
Each five-slice stack equals the sum of its slices to within 2\,ms.
We did not time two-slice stacks directly.
Summing the nearest and farthest slices of each run gives an estimate of 1.16 $\pm$ 0.02\,s, 60\% less than a five-slice stack.

\section{Additional Qualitative Results}
\label{sec:supp_qual}
\begin{figure*}[p]
\centering
\includegraphics[width=\linewidth]{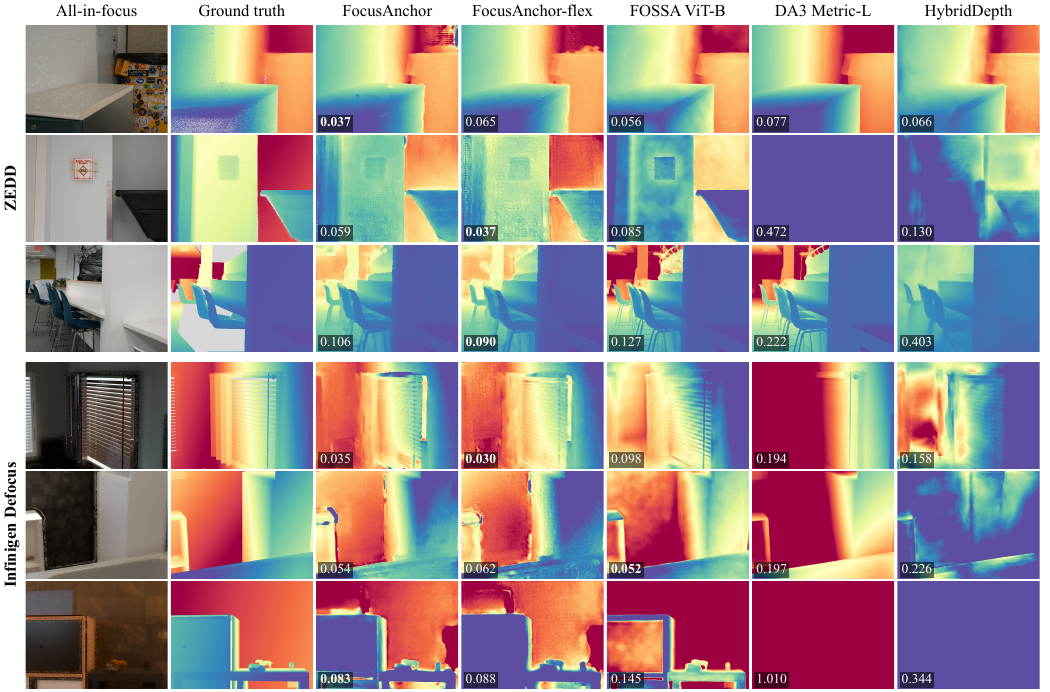}
\caption{\textbf{Typical zero-shot results on ZEDD and Infinigen Defocus.} Rows are the images at the 25th, 50th and 75th percentile of the per-image AbsRel of \method. Each row has one color range, from the 2nd to the 98th percentile of the ground truth, with blue for near and red for far. A prediction whose metric scale lies beyond this range fills the map with a single color. \method and \method-flex stay close to the ground-truth colors. Gray marks pixels without ground truth. Numbers are per-image AbsRel, with the best in bold. On Infinigen Defocus, \cref{tab:zeroshot} takes the FOSSA and HybridDepth numbers from \cite{zuo2026fossa}, while the maps here are our runs of the released models.}
\label{fig:supp_qual_a}
\end{figure*}

\begin{figure*}[p]
\centering
\includegraphics[width=\linewidth]{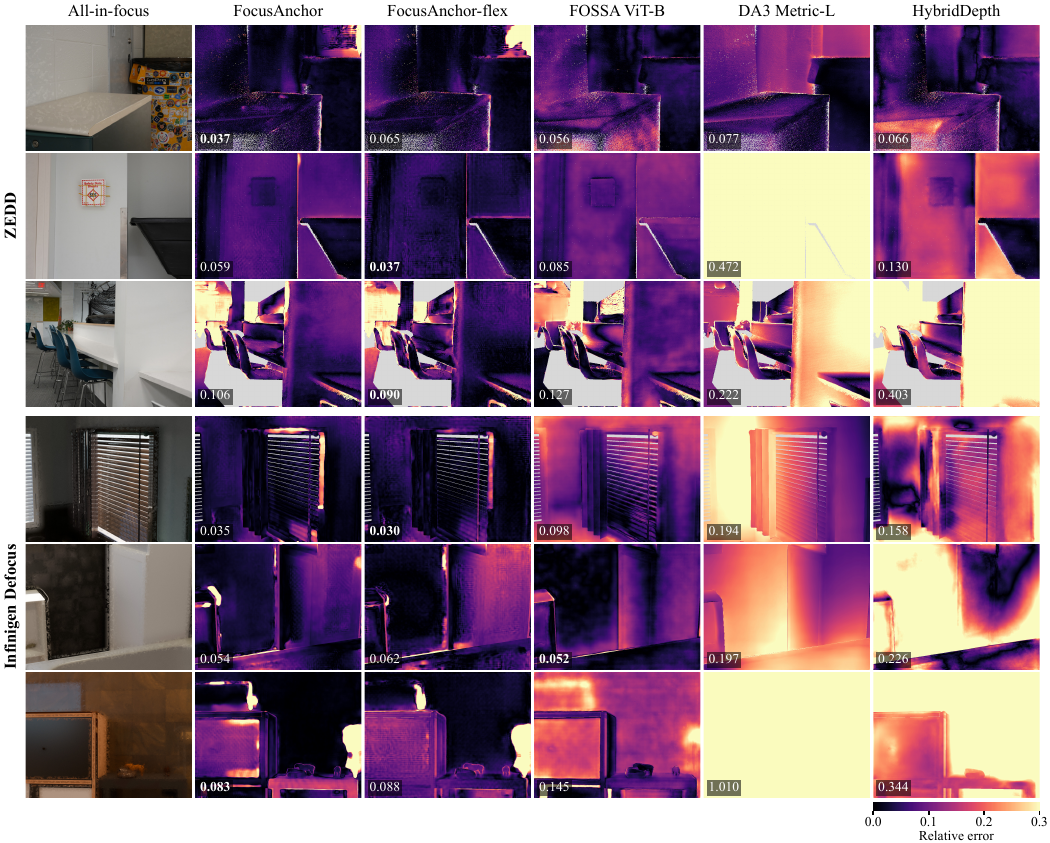}
\caption{\textbf{Relative error on ZEDD and Infinigen Defocus} for the predictions of \cref{fig:supp_qual_a}. Each pixel shows $|\hat{d}-d|/d$ for predicted depth $\hat{d}$ and ground truth $d$, from black (0) to light yellow (0.3 or more). Gray marks pixels without ground truth, and each number is the mean error over the remaining pixels (per-image AbsRel). The error of \method and \method-flex sits in thin lines at depth edges and in a few small regions. The error of the other methods often covers whole surfaces, which is the pattern of a wrong scale.}
\label{fig:supp_qual_a_err}
\end{figure*}

\begin{figure*}[p]
\centering
\includegraphics[width=\linewidth]{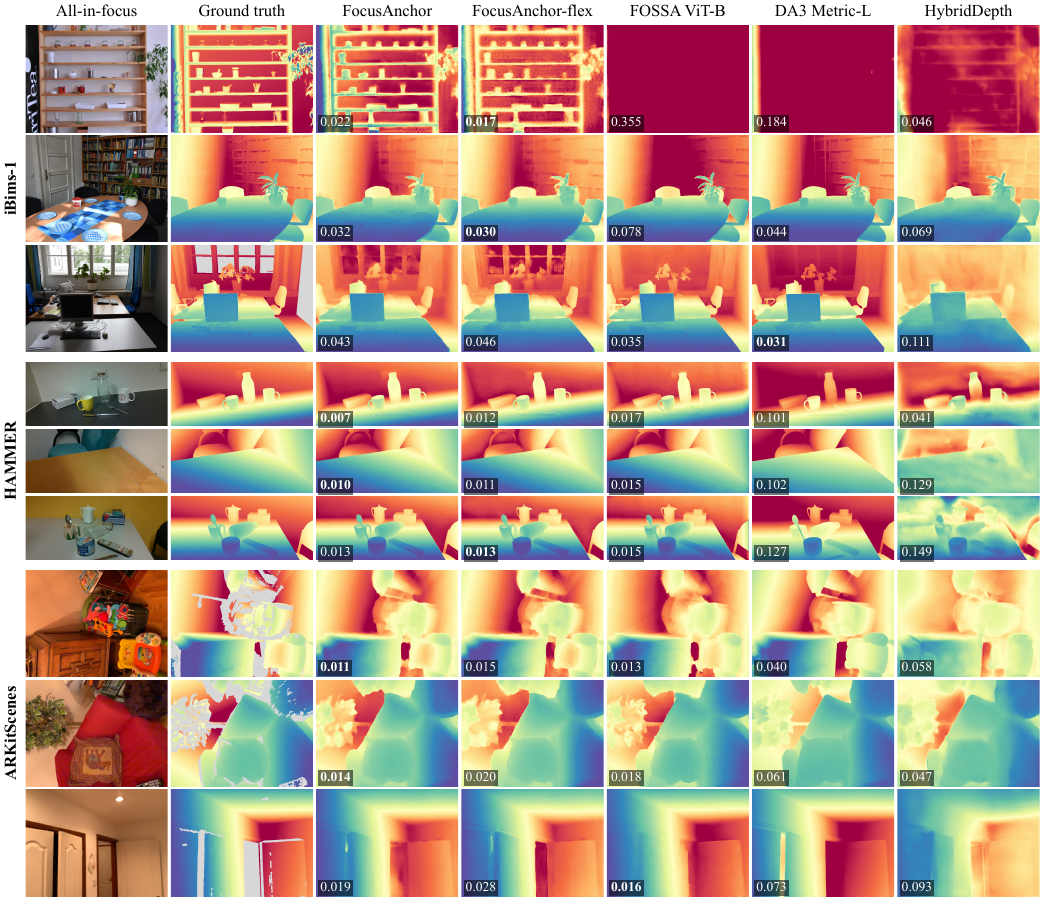}
\caption{\textbf{Typical zero-shot results on iBims, HAMMER and ARKitScenes.} Focal stacks are rendered as in \cref{tab:zeroshot}. Rows, color ranges, gray pixels and numbers are as in \cref{fig:supp_qual_a}. In the first iBims row, FOSSA ViT-B and DA3 Metric-L place the shelf beyond the far end of the color range, so their maps show a single color. On HAMMER, DA3 Metric-L keeps the layout of the scene at a shifted scale, and HybridDepth blurs object shapes. \method and \method-flex keep both the scale and the shapes.}
\label{fig:supp_qual_b}
\end{figure*}

\begin{figure*}[p]
\centering
\includegraphics[width=\linewidth]{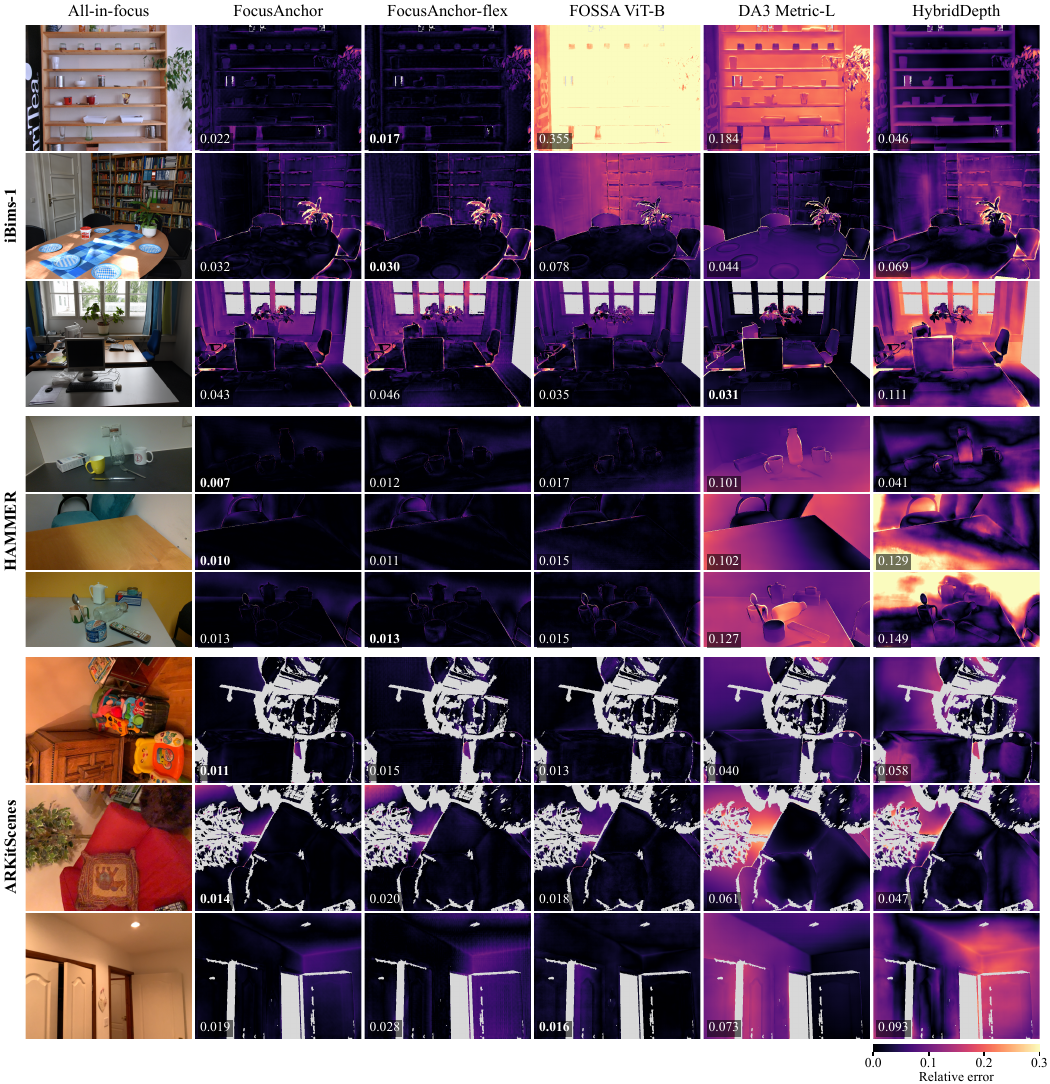}
\caption{\textbf{Relative error on iBims, HAMMER and ARKitScenes} for the predictions of \cref{fig:supp_qual_b}. Pixels, colors, gray and numbers are as in \cref{fig:supp_qual_a_err}. The error of \method and \method-flex lies mostly at depth edges and thin structures. On HAMMER, the error of DA3 Metric-L covers whole surfaces, as expected from a shifted scale, and the error of HybridDepth forms large uneven patches.}
\label{fig:supp_qual_b_err}
\end{figure*}

\Cref{fig:supp_qual_a,fig:supp_qual_b} show three scenes from each zero-shot benchmark of \cref{tab:zeroshot}, and \cref{fig:supp_qual_a_err,fig:supp_qual_b_err} show the relative error of the same predictions.
The scenes in \cref{fig:qual} were picked by hand. Here, for each benchmark, we take the images at the 25th, 50th and 75th percentile of the per-image AbsRel of \method, with at most one image per scene or video. The rows therefore show typical cases, not the best ones.
Every method gets the same inputs as in \cref{tab:zeroshot}.
\Cref{fig:qual_error} shows the per-pixel error for the hand-picked scenes of \cref{fig:qual}.

\method or \method-flex has the lowest AbsRel on 12 of the 15 rows.
The focus planes give both of our models the metric scale. On every row, the median ratio of our depth to the ground truth lies within 7\% of one.
DA3 Metric-L has to infer the scale from the image alone, and its maps often have the right shape at the wrong scale. One scale factor per image removes most of its error on 8 of the 15 rows.
On HAMMER and ARKitScenes, the same rescaling changes the error of HybridDepth by at most 11\%, so its error there lies in the shape of the scene rather than in its scale.
Part of our remaining error lies beyond the farthest focus plane, where blur hardly changes with depth. In the third ZEDD row, the 17\% of pixels beyond the 8.1\,m plane carry 39\% of the error of \method.

\section{Slice Choice in the Stack-Size Study}
\label{sec:supp_stack}
ZEDD provides nine focus settings per scene.
\Cref{tab:stack} uses settings $\{1,9\}$, $\{1,5,9\}$, $\{1,4,6,9\}$ and $\{1,3,5,7,9\}$ for two to five slices, so every stack spans 0.82 to 8.10\,m.
The all-in-focus image is fused from the selected slices only.
FOSSA~\cite{zuo2026fossa} reports a stack-size study on the same split without stating which settings it uses.
Its $\delta_{1.25}$ with five and four slices (0.919 and 0.865, read from its Fig.~5c) agrees with our runs of its released ViT-B model (0.919 and 0.857).
Its values with two and three slices (0.702 and 0.888) are higher than ours (0.421 and 0.680), so the difference likely lies in the slice choice.
\method-flex (0.855, 0.899, 0.876 and 0.920 for two to five slices) is above FOSSA's reported values at every size.
Four slices are the only case that cannot be evenly spaced among the nine settings, and both \method-flex and FOSSA's own study dip there.